\documentclass[PHM, 2025]{PHMSociety}
\usepackage{graphicx}
\usepackage{amsmath}
\usepackage{tikz}
\usepackage{svg}
\makeatletter
\providecommand\@xviiipt{17.28}
\providecommand\@xxiiipt{22.2}
\makeatother
\usepackage{enumitem}
\usepackage{booktabs}
\usepackage{longtable}
\usepackage{mdframed}
\usepackage{algpseudocode}
\usepackage{algorithm2e}
\usepackage{booktabs, tabularx}
\usepackage{siunitx}
\usepackage{caption,subcaption}
\newcolumntype{Y}{>{\raggedright\arraybackslash}X}
\usetikzlibrary{positioning, shapes.multipart, shapes.geometric, shapes, arrows.meta, fit, calc}
\usepackage[most]{tcolorbox}
\usepackage{fontawesome5} 
\tcbset{
  colback=black!2,
  colframe=black!35,
  boxrule=0.5pt,
  arc=2mm,
  left=2.5mm,right=2.5mm,top=1.5mm,bottom=1.5mm,
  enhanced, breakable
}
\newtcolorbox{promptbox}[1]{title={#1}, fonttitle=\bfseries, parbox=false}

\begin{document}

\title{Cleaning Maintenance Logs with LLM Agents for Improved Predictive Maintenance }

\author{%
	Valeriu Dimidov\authorNumber{}, Faisal Hawlader\authorNumber{}, Sasan Jafarnejad\authorNumber{} and Raphaël Frank\authorNumber{}
}

\address{
	\affiliation{{}}{Interdisciplinary Centre for Security, Reliability and Trust (SnT)\\
    University of Luxembourg \\ 
    29 Avenue J.F. Kennedy L-1855, Luxembourg}{ 
		{\email{firstname.lastname@uni.lu}}\\ 
		} 
	\tabularnewline 
}

\maketitle

\begin{abstract}
Economic constraints, limited availability of datasets for reproducibility and shortages of specialized expertise have long been recognized as key challenges to the adoption and advancement of predictive maintenance (PdM) in the automotive sector. 
Recent progress in large language models (LLMs) presents an opportunity to overcome these barriers and speed up the transition of PdM from research to industrial practice. 
Under these conditions, we explore the potential of LLM-based agents to support PdM cleaning pipelines. 
Specifically, we focus on maintenance logs, a critical data source for training well-performing machine learning (ML) models, but one often affected by errors such as typos, missing fields, near-duplicate entries, and incorrect dates. 
We evaluate LLM agents on cleaning tasks involving six distinct types of noise. Our findings show that LLMs are effective at handling generic cleaning tasks and offer a promising foundation for future industrial applications. While domain-specific errors remain challenging, these results highlight the potential for further improvements through specialized training and enhanced agentic capabilities.
\end{abstract}

\section{Introduction}

Industrial data-driven PdM initiatives involving automotive vehicles often span several years due to the rarity of failures and the need to accumulate extensive sensor data. 
In their early stages, such projects primarily consist of passively collecting operational data and maintenance logs. 
However, the data acquisition process is typically neither monitored by humans nor supported by automated mechanisms to validate the correctness of raw data. 
Consequently, the resulting datasets are frequently noisy and require extensive cleaning before being used in downstream tasks.

These data quality issues are well-documented in the PdM literature. 
For instance, the authors of \cite{prytz2015} report common problems in truck maintenance logs, such as missing entries, manual input errors, and low fault resolution. 
They note that maintenance logs were not originally designed for data mining purposes and argue that such limitations introduce substantial label noise into predictive models. 
Along the same lines, the authors of \cite{delmoral2022} emphasize that real-world repair logs related to hospital sterilizers often contain uncertain dates, undocumented interventions, and records that do not reflect actual failures but rather preventive maintenance activities. 
In addition, in an ad hoc study aimed at identifying key data quality pitfalls that prevent Finnish multinational OEMs from providing effective after-sales maintenance services, the authors of \cite{madhikermi_key_2017} highlight that technicians frequently omit critical fields such as component codes, reason codes, and action codes. 
These omissions severely hinder root cause analysis, failure prediction, and the training of reliable models.

Currently, data cleaning activities are carried out at the end of the monitoring phase of a PdM project. They follow an iterative error-detection and error-repair cycle, relying on pipelines implemented through scripts and software tools. Nevertheless, the process remains partly manual, time-consuming, and error-prone, often failing to eliminate all sources of inconsistency. As a result, some records cannot be repaired and are discarded, while others are only partially corrected, leaving residual noise that ultimately degrades the performance of predictive models.

This challenge motivates the exploration of novel AI-driven approaches, such as LLM-based agents, to enable more efficient and reliable data curation in PdM. In particular, LLMs offer the potential to shift the cleaning paradigm from batch-oriented processing to stream-based, real-time correction, allowing agents to detect and repair errors as records are ingested. This study represents an initial step toward a broader investigation into whether LLM-based maintenance log cleaning can outperform traditional approaches. By benchmarking LLM agents across diverse noise types, we aim to assess their strengths, limitations, and suitability for industrial deployment. To support this investigation, our contributions are fourfold:
\begin{enumerate}
    \item Define a taxonomy of common noise patterns specific to automotive PdM data.
    \item Propose an open source framework for generating synthetic logs with controlled noise. The framework, named \textit{AgenticPdmDataCleaner}, is publicly available \footnote{\scalebox{0.95}{\url{https://github.com/sntubix/agentic-pdm-log-cleaning}}}.
    \item Benchmark multiple LLMs on cleaning tasks.
    \item Analyze error types and limitations, providing guidance for adapting LLM agents to industrial PdM settings.
\end{enumerate}

The remainder of this manuscript is structured as follows. 
Section 2 reviews related work on data cleaning techniques, ranging from classical rule-based and probabilistic approaches to recent LLM-driven frameworks. 
Section 3 presents our framework for synthetic fleet data generation, including the system model, data sources, and controlled noise injection mechanisms designed to reproduce real-world inconsistencies in maintenance logs. 
Section 4 introduces our methodology, detailing the LLM-based agent framework, task definitions, and evaluation setup. 
Section 5 provides information about the benchmarking configuration and the metrics used to evaluate the experiments. 
Section 6 reports the benchmarking results, while Section 7 discusses the scientific and industrial implications of our findings, emphasizing both the strengths and limitations of the proposed approach. Finally, Section 8 concludes the paper and outlines directions for future research.

\section{Related Work}
Classical data cleaning approaches rely on rule-based validation, statistical profiling, and integrity constraints to detect and correct inconsistencies \cite{fan_foundations_2012, ilyas_trends_2015}. 
To overcome their limitations, \cite{chu_katara_2015} proposed Katara as a system that leverages knowledge bases and crowdsourcing. 
The system maps table semantics to external knowledge, validates uncertain cases with human input, and suggests top-k repairs for erroneous tuples, thereby combining automated reasoning with selective crowd involvement. 
HoloClean extends this line of research by introducing a probabilistic inference framework that unifies signals from integrity constraints, external data, and statistical co-occurrence patterns to perform holistic data repairing, significantly improving repair quality compared to isolated methods \cite{rekatsinas_holoclean_2017}.

Subsequent works, increasingly based on ML, divided the data cleaning task into error detection and error correction. 
HoloDetect addresses detection by framing it as a few-shot learning problem \cite{heidari_holodetect_2019}, while Raha automates the configuration of multiple detection strategies to minimize manual intervention \cite{mahdavi_raha_2019}. 
For correction, Baran provides a unified framework that ensembles candidate repairs through semi-supervised and transfer learning \cite{mahdavi_baran_2020}.

More recently, large language models have been explored as general-purpose engines for structured data curation. 
\cite{narayan_can_2022} demonstrated that foundation models like GPT-3 can be adapted to entity matching, error detection, schema matching, and imputation via prompting, achieving competitive performance with minimal supervision. 
\cite{zhang_large_2024} extended this view by benchmarking GPT-3.5, GPT-4, and GPT-4o as data preprocessors across multiple tasks, showing that LLMs can rival specialized baselines when guided by prompt engineering techniques such as zero/few-shot conditioning, contextualization, and batch prompting. 
Beyond static prompting, \cite{bendinelli_exploring_2025} introduced a benchmark where LLM agents iteratively clean intentionally corrupted datasets through tool calls and performance feedback, correcting simple row-level anomalies but struggling with distributional shifts and contextual errors. 
Finally, \cite{qi_cleanagent_2025} proposed CleanAgent, which integrates declarative APIs with LLM-based agents to automate data standardization tasks such as date and address formatting.

Despite these advances, the potential of LLM-based agents to curate domain-specific PdM logs remains largely unexplored, leaving an open question about their applicability to industrial environments.

\begin{figure*}[h]
    \centering
    \includegraphics[width=0.65\textwidth]{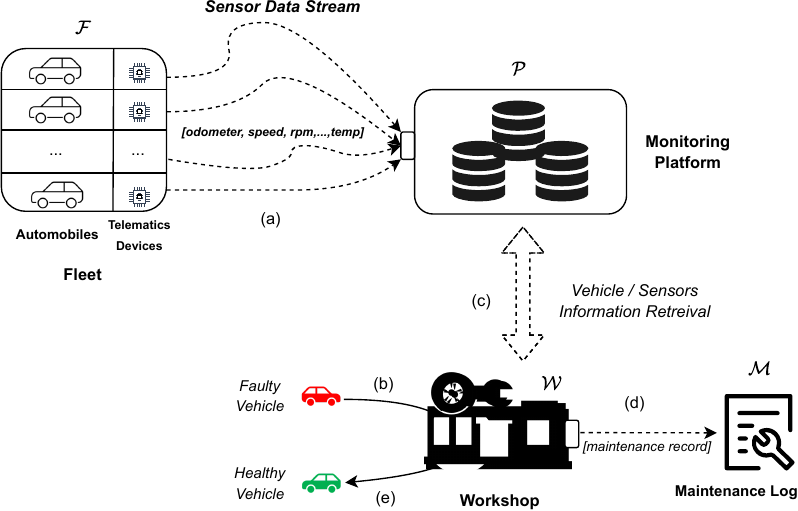}
    \caption{Fleet Monitoring, Repair, and Maintenance Logging Process.}
    \label{fig:system_model}
\end{figure*}

\section{Framework for Synthetic Fleet Data Generation with Noise Injection}

In the automotive sector, the release of production data is rare. 
This is primarily due to privacy concerns and the reluctance of OEMs and workshops to disclose details about equipment reliability or internal processes. 
To address this limitation, we developed a synthetic data generation framework that serves as a proxy for the real-world scenarios we aim to investigate. 
The log schema used in our framework is a simplified adaptation of the ontology proposed by \cite{woods_ontology_2024}. 
Nevertheless, for the scope of this study, the generated logs are sufficient.

The framework supports the generation of synthetic fleet data in both tabular and time-series formats. 
It includes mechanisms for controlled noise injection into tabular data, enabling systematic evaluation of an agent’s ability to detect and repair inconsistencies when correlating multiple tables or linking tabular records with time-series signals. 
In the current implementation, the fleet registry and time-series data are generated without noise and serve as clean reference signals. 
These can be leveraged by agents to infer or correct corrupted entries in the maintenance logs.

A synthetic data generator provides several benefits. 
It enables the creation of diverse datasets for studying the statistical significance of proposed methodologies and can be safely ingested by LLMs without raising privacy concerns. 
Furthermore, given that LLMs tend to memorize benchmark datasets, synthetic generators provide a means to overcome this issue by allowing benchmarking on datasets with similar statistical distributions but novel instances.

\begin{figure*}[t]
  \centering
  \begingroup
  \setlength{\tabcolsep}{3.5pt}
  \renewcommand{\arraystretch}{0.92}
  \footnotesize

\makebox[\textwidth][c]{%
  \begin{subtable}[t]{0.62\textwidth}
    \centering
    \caption{Fleet Registry}
    \label{tab:device_registry_excerpt}
    \begin{tabularx}{\textwidth}{@{}l l l l >{\ttfamily}X@{}}
      \toprule
       & \textbf{device\_id} & \textbf{name} & \textbf{license\_plate} & \textbf{VIN} \\
      \midrule
      0 & b754A & (b754A) & AH4657 & KT6HA8KW6LWZD5747 \\
      1 & b242F & (b242F) & CT9935 & 0T1UNZHC09032KBLY \\
      2 & b189E & (b189E) & KA4582 & RKR3PC0K6HVW3ZSWA \\
      3 & b338E & (b338E) & PO9928 & N6NGFAHS53H7R4C44 \\
      \bottomrule
    \end{tabularx}
  \end{subtable}%
}
  \hfill
  \begin{minipage}[t]{0.485\textwidth}\vspace{0pt}\end{minipage}

  \par\medskip

  \begin{subtable}[t]{0.485\textwidth}
    \centering
    \caption{Sensor Table}
    \label{tab:b787F_odometer}
    \begin{tabularx}{\textwidth}{@{}l l l S[table-format=6] S[table-format=3]@{}}
      \toprule
       & \textbf{device\_id} & \textbf{date} & \textbf{odometer} & \textbf{km\_traveled} \\
      \midrule
      0 & b338E & 2022-06-14 & 358156 & 196 \\
      1 & b338E & 2022-06-15 & 358257 & 100 \\
      2 & b338E & 2022-06-16 & 358257 & 0 \\
      3 & b338E & 2022-06-17 & 358257 & 0 \\
      4 & b338E & 2022-06-18 & 358257 & 0 \\
      5 & b338E & 2022-06-19 & 358257 & 0 \\
      6 & b338E & 2022-06-20 & 358257 & 0 \\
      7 & b338E & 2022-06-21 & 358365 & 108 \\
      8 & b338E & 2022-06-22 & 358556 & 190 \\
      \bottomrule
    \end{tabularx}
  \end{subtable}\hfill
  \begin{subtable}[t]{0.485\textwidth}
    \centering
    \caption{Service Operations Catalog}
    \label{tab:service_ops_excerpt}
    \begin{tabularx}{\textwidth}{@{}l Y Y Y Y@{}}
      \toprule
       & \textbf{System} & \textbf{Subsystem} & \textbf{Component} & \textbf{Activity} \\
      \midrule
      0 & Powertrain & Engine & Cylinder Head & Repair \\
      1 & Brake System & Hydraulic Brake & Brake Pads & Replace \\
      2 & HVAC & Air Conditioning & Compressor & Repair \\
      3 & Steering & Rack and Pinion & Steering Rack & Replace \\
      4 & HVAC & Cooling & Coolant Pump & Replace \\
      \bottomrule
    \end{tabularx}
  \end{subtable}

  \par\medskip

\begin{subtable}[t]{\textwidth}
    \centering
    \scriptsize
    \caption{Maintenance Log}
    \label{tab:maint_log_excerpt}
    \begin{tabularx}{\textwidth}{@{}c l l l l l l l >{\raggedright\arraybackslash}X@{}}
      \toprule
      \textbf{wo\_num} & \textbf{start\_date} & \textbf{end\_date} & \textbf{license\_plate} &
      \textbf{system} & \textbf{subsystem} & \textbf{component} & \textbf{activity} & \textbf{work\_description} \\
      \midrule
      WO129 & 2021-05-03 & 2021-05-07 & AH4657 & Powertrain & Engine & Cylinder Head & Repair &
      Repaired cylinder head. \\
      WO827 & 2021-01-02 & 2021-01-06 & CT9935 & Brake System & Hydraulic Brake & Brake Pads & Replace &
      Replaced brake pads. \\
      WO329 & 2021-08-31 & 2021-09-04 & KA4582 & HVAC & Air Conditioning & Compressor & Repair &
      Repaired air conditioning compressor. \\
      WO679 & 2022-06-16 & 2022-06-21 & PO9928 & Steering & Rack and Pinion & Steering Rack & Replace &
      Replaced steering rack. \\
      \bottomrule
    \end{tabularx}
\end{subtable}

  \endgroup
  \caption{Clean data excerpts: (a) Fleet Registry, (b) Sensor Table, (c) Service Operations Catalog, (d) Maintenance Log.}
  \label{fig:data_excerpts}
\end{figure*}
\begin{figure*}
  \centering
  \begingroup
  \setlength{\tabcolsep}{3.5pt}
  \renewcommand{\arraystretch}{0.92}
  \footnotesize

  \begin{minipage}[t]{0.485\textwidth}\vspace{0pt}\end{minipage}

  \par\medskip

\begin{subtable}[t]{\textwidth}
    \centering
    \scriptsize
    \caption{Maintenance Log}
    \label{tab:noisy_maint_log}
    \begin{tabularx}{\textwidth}{@{}c l l l l l l l l >{\raggedright\arraybackslash}X@{}}
      \toprule
      \textbf{wo\_num} & \textbf{start\_date} & \textbf{end\_date} & \textbf{license\_plate} &
      \textbf{system} & \textbf{subsystem} & \textbf{component} & \textbf{activity} & \textbf{work\_description} & \textbf{label} \\
      \midrule
      WO129 & 2021-05-03 & 2021-05-07 & \textit{(b754A)} & Powertrain & Engine & Cylinder Head & Repair &
      Repaired cylinder head. & M1 \\
      WO827 & 2021-01-02 & 2021-01-06 & CT9935 & \textit{Brake Sysem} & Hydraulic Brake & Brake Pads & Replace &
      Replaced brake pads. & M3 \\
      WO329 & 2021-08-31 & 2021-09-04 & KA4582 & HVAC & Air Conditioning &  & Repair &
      Repaired air conditioning .... & M4 \\
      WO679 & 2022-06-16 & \textit{2022-06-21} & PO9928 & Steering & Rack and Pinion & Steering Rack & Replace &
      Replaced steering rack. & M6 \\
      WO333 & 2022-06-16 & 2022-06-21 & TEST & - & - & - & Test &
      Testing the IT system. & M5 \\
      WO429 & 2021-08-31 & 2021-09-04 & WI0000 & HVAC & Cooling & Coolant Pump & Replace &
      Replaced worn-out ... & M2 \\
      \bottomrule
    \end{tabularx}
\end{subtable}

  \endgroup
  \caption{Noisy Maintenance Log.}
  \label{fig:data_excerpts1}
\end{figure*}
\subsection{System Model}

The overall architecture of the fleet monitoring and maintenance logging process is depicted in Figure~\ref{fig:system_model}. 
The fleet \( \mathcal{F} = \{v_1, v_2, \ldots, v_N\} \) consists of \( N \) vehicles, each equipped with telematics devices that stream telemetry to a central monitoring platform \( \mathcal{P}\).
When a failure is detected, the affected vehicle is sent to a workshop \( \mathcal{W} \) for repair, where technicians consult the platform \( \mathcal{P}\) to retrieve diagnostic data and vehicle information. 
Once the intervention is completed, the maintenance activity is recorded in the maintenance log \( \mathcal{M} \), which contains both administrative details—such as identifiers, dates, and vehicle references—and technical details describing the affected system, subsystem, and component, the activity performed, and additional contextual metadata. 
A complete schema of \(\mathcal{M}\) is provided in Section~\ref{sec:maintenance_log}.

\subsection*{Modeling Assumptions}

To isolate the impact of data quality issues in the maintenance logs, we assume the following simplifications:
\begin{enumerate}[label=A\arabic*]
    \item \textbf{Fleet Homogeneity} — all vehicles share the same make, model, and component specifications.
    \item \textbf{Fleet Staticity} — no vehicles are added to or removed from the fleet during the monitoring period.
    \item \textbf{Uniform Operational Profile} — all vehicles are assumed to operate under the same usage patterns and duty cycles.
    \item \textbf{Single Maintenance Event Constraint} — each vehicle experiences at most one maintenance event.
    \item \textbf{Centralized Maintenance Facility} — all repairs are performed at a single facility using a standardized maintenance log schema.
    \item \textbf{Corrective Repairs} — all the maintenance records are related to corrective maintenance activities
\end{enumerate}

Although these assumptions simplify the naturally diverse and evolving characteristics of real-world fleet management, and by extension, the preprocessing of maintenance records, they enable the construction of a controlled benchmarking dataset. 
Future developments will aim to relax these constraints to better reflect real-world conditions.

\subsection{Data Sources}
\label{sec:data_sources}
The process described in the previous section gives rise to four data sources that capture different aspects of the fleet monitoring and maintenance workflow: \textit{Fleet Registry}, \textit{Sensor Data}, \textit{Service Operations Catalog}, and \textit{Maintenance Log}. 
Together, they provide the foundation for validating, cleaning and transforming maintenance records. 

\subsubsection{Fleet Registry}
\label{sec:fleet_registry}

The fleet registry \( \mathcal{F} \) stores the master records of all vehicles in the monitored fleet. 
Each entry corresponds to a unique device installed on a vehicle and includes key identifiers such as the \texttt{device\_id}, \texttt{name}, \texttt{license\_plate}, and \texttt{VIN}. 

This registry serves as a reference for linking vehicle identifiers across sensor data, maintenance logs, and other datasets.

\subsubsection{Sensor Data}
\label{sec:sensor_data}

The sensor dataset \( \mathcal{S} \) contains time-stamped operational measurements collected from onboard devices. 
In the current implementation, it includes a single signal, namely the \texttt{odometer\_km} reading for each vehicle, indexed by the unique \texttt{device\_id} and a \texttt{date} field. 
The \texttt{odometer\_km} values represent the cumulative distance traveled by the vehicle at the given date, providing a monotonically increasing metric of vehicle usage. 
The linkage of \texttt{device\_id} to the fleet registry ensures consistent association between sensor readings and the corresponding vehicle metadata.

\subsubsection{Service Operations Catalog}
Each intervention performed during a workshop visit can be mapped to a predefined taxonomy, referred to as the service operations catalog. 
In our service operations catalog, each workshop intervention is organized along a three-tier hierarchy. 
At the top level is the system, a broad functional domain of the vehicle such as Powertrain, Chassis, or Electrical. 
Within each system, we distinguish subsystems that narrow the focus of the task (e.g., the Engine or Transmission within the Powertrain system). 
Finally, the most specific category is the component, which identifies the exact part addressed during the intervention. 
Beyond this hierarchical classification, every record also specifies the type of activity, such as replacement or repair. 

The service operations catalog used in this work spans 10 systems, 26 subsystems, and 34 components across 142 entries. 
Unlike the other data sources, it is static and was constructed manually.

\subsubsection{Maintenance Log}
\label{sec:maintenance_log}

The maintenance log \( \mathcal{M} \) contains structured fields that capture the administrative and technical aspects of a maintenance intervention. 
Administrative fields include identifiers such as the work order number (\texttt{wo\_num}) as well as temporal information (\texttt{start\_date}, \texttt{end\_date}) and the \texttt{license\_plate} used to link the intervention to a specific vehicle. 
Technical fields specify the scope of the intervention, from the high-level \texttt{system} down to the \texttt{subsystem} and individual \texttt{component}, along with the performed \texttt{activity} and its textual \texttt{work\_description}. 
Additional metadata provide contextual information for subsequent analyses: \texttt{work\_order\_type} indicates whether the intervention was corrective, preventive, or predictive.

\subsection{Noise Injection Framework}

To realistically evaluate the ability of LLM-based agents to clean maintenance records, we introduce a noise injection framework that systematically corrupts otherwise clean synthetic logs. 
This framework is grounded in a taxonomy of common errors observed in industrial datasets—such as identifier misalignments, missing values, and incorrect dates—and provides mechanisms to reproduce them in a controlled manner.
By generating paired clean and noisy datasets with configurable proportions of each noise type, the framework enables reproducible benchmarking and fine-grained analysis of model performance under diverse data quality challenges.

\subsubsection{Noise Taxonomy}
\label{sec:noise_maintenance_log}

In real-world PdM deployments, maintenance logs rarely conform perfectly to their intended schema. 
Noise can arise from human error, inconsistent data entry practices, or incomplete integration between workshop and fleet monitoring systems. 
For the purposes of our study, we denote the absence of noise as M0 and introduce six additional noise types:
\begin{enumerate}[label=M\arabic*]
    
  \item \textbf{Vehicle identifier misalignment} – The default vehicle identifier field \texttt{license\_plate} is replaced with \texttt{device\_id}, \texttt{name}, or \texttt{VIN}, breaking the linkage between maintenance records and the fleet registry. 
  As shown in Figure~\ref{tab:noisy_maint_log}, record \texttt{WO129} has (b754A) as \texttt{license\_plate}, which is a device name in Figure~\ref{tab:device_registry_excerpt}.

  \item \textbf{Out-of-fleet vehicles} – Records reference valid plates belonging to vehicles outside the monitored fleet \( \mathcal{F} \), introducing exogenous observations. 
  For example, Figure~\ref{tab:noisy_maint_log}, record \texttt{WO429} lists \texttt{WI0000} as \texttt{license\_plate}, which is absent from the Fleet Registry (Figure~\ref{tab:device_registry_excerpt}).

  \item \textbf{Invalid values} – Categorical fields (\texttt{system}, \texttt{subsystem}, \texttt{component}, \texttt{activity}) contain tokens outside the controlled vocabulary (typos or non-standard labels). 
  For instance, in Figure~\ref{tab:noisy_maint_log} the record \texttt{WO827} has a typo in the field \texttt{system}.

  \item \textbf{Missing values} – One categorical field is left empty, yielding structurally missing information. 
  In Figure~\ref{tab:noisy_maint_log}, record \texttt{WO329} contains an empty \texttt{component} field.

  \item \textbf{Digital system test} – Entries document installation, calibration, or testing of the monitoring system rather than vehicle maintenance interventions and should be segregated. 
Considering Figure~\ref{tab:noisy_maint_log}, the record \texttt{WO333} has numerous fields labeled as TEST.

  \item \textbf{Wrong end dates} – The \texttt{end\_date} is inconsistent with the intervention timeline inferred from operational signals. 
  Specifically, record \texttt{WO679} reports \texttt{end\_date}=\textit{2022-06-21}, which conflicts with the usage pattern for \texttt{b338E} in Figure~\ref{tab:b787F_odometer}.

\end{enumerate}

These categories represent the primary types of noise observed in a real-world log and fleet environment provided by Grupo\textless A\textgreater\footnote{\url{https://www.grupoa.co/}}, a Colombian multinational operating across various industrial sectors including automotive equipment manufacturing, machinery, and mining. The dataset was collected between 2021 and 2024. It includes structured fields such as vehicle plate number, work order ID, component name, action code, mechanic information, timestamps, odometer readings, cost of maintenance, and system/subsystem classification. This rich schema enabled the identification and categorization of real-world noise patterns in fleet maintenance logs and was instrumental in the development and validation of the proposed methods.

\subsubsection{Noise Injection Mechanisms}

\subsubsection*{Definitions}
\begin{itemize}
    \item \( N \): Total number of entries to generate.
    \item \( T = \{ t_1, t_2, \ldots, t_K \} \): Set of noise types 
    \item \( \pi = (\pi_1, \pi_2, \ldots, \pi_K) \): Proportions of each noise type, with \( \sum_{k=1}^K \pi_k = 1 \).
    \item \( \mathcal{D}_k \subset \mathcal{D} \): Subset of entries of type \( t_k \), with \( |\mathcal{D}_k| = \pi_k N \).
\end{itemize}

Each noise type $t_k \in T$ is associated with a noise generator $S_k$, which defines how corrupted or clean entries are created. 
Regardless of the noise category, each generator yields two aligned datasets: the clean records $\mathcal{E}$ and the noisy records $\mathcal{E}'$.  
The framework is designed to be fully configurable, allowing the user to adjust the proportions $\pi_k$ of each noise type to match specific experimental setups. 
Moreover, the noise taxonomy is extensible: new noise types can be seamlessly incorporated by defining additional generators and integrating them into $T$. 
This design enables the creation of diverse and realistic noise patterns, supporting both controlled experiments and exploratory evaluations.

\begin{itemize}
    \item \textbf{Absence of Noise}
    \[
    \mathcal{S}_k : \mathcal{E} \rightarrow \mathcal{E}
    \quad 
    \]
    This generator returns the input entry unaltered. 
    It is used to generate noise-free records and corresponds to \( t_k = t_0 \), the special case representing absence of noise.
    
    \item \textbf{Corruptive Noise}
    \[
    \mathcal{S}_k : \mathcal{E} \rightarrow \mathcal{E}' 
    \quad
    \]
    The generator receives a clean entry \( \mathcal{E} \) and applies a transformation to produce a corrupted version \( \mathcal{E}' \).
    
    \item \textbf{Generative Noise}
    \[
    \mathcal{S}_k : \emptyset \rightarrow \mathcal{E}' 
    \quad
    \]
    The generator does not rely on an existing clean entry but generates corrupted entries from scratch.
\end{itemize}

\begin{table}[ht]
\centering
\caption{Classification of noise types}
\label{tab:noise_types}
\begin{tabular}{|c|l|c|}
\hline
\textbf{ID} & \textbf{Name} & \textbf{Noise Type} \\
\hline
M1 & Vehicle Id misalignment & Corruptive \\
M2 & Out-of-fleet vehicles & Generative \\
M3 & Invalid values & Corruptive \\
M4 & Missing values & Corruptive \\
M5 & Digital system test & Generative \\
M6 & Wrong end dates & Corruptive \\
\hline
\end{tabular}
\end{table}

Table ~\ref{tab:noise_types} maps each implemented noise type to the corresponding class in our noise taxonomy.

\subsubsection{Fleet Data Generation}
\label{sec:fleet_data_generation}
The fleet data generation process follows a two-step approach. 
First, we create a clean dataset that integrates information from the fleet registry, service operations catalog, and sensor signals to produce consistent maintenance records. 
This clean version serves as the ground truth. 
In the second step, we regenerate the maintenance log by injecting controlled noise according to predefined categories, while keeping the other data sources unchanged. 
The result is a pair of aligned datasets—clean and noisy—that enable systematic benchmarking of LLM-based agents under realistic data quality challenges.

\subsubsection*{Clean Data Sources}
The process of clean fleet data generation depends on three key parameters: the time interval during which the fleet is monitored, the country of registration, and a static service operations catalog. 

The generation begins with the creation of the \textit{device\_table}, which lists all vehicles in the simulated fleet. 
For each vehicle, a unique internal identifier is assigned, following a predefined pattern (e.g., \texttt{b742C}), along with a license plate generated according to the country of registration format. 
Each entry also includes a globally unique vehicle identification number (VIN) and a display name, which by default is the device id enclosed in parentheses. 
The number of entries in this table is determined by the expected number of maintenance records, as defined in Assumption~A1.

Once the fleet registry is established, the \textit{maintenance table} is generated.  
Currently, all noise generators in the framework inherit from a common noise-free schema \(\mathcal{E}\), which defines the structure and content of clean maintenance records.  
Under this schema, each vehicle is associated with an activity randomly drawn from the service operations of the workshop. 
The start date of each record is selected uniformly at random within the monitoring period, and the end date is set between four and seven days later.  
The textual \texttt{work\_description} is produced by a large language model instructed to include both the component and activity terms in a concise, technician-style note.

The final stage involves producing the \textit{sensor time-series}, which records daily odometer readings for each vehicle across the monitoring window. 
The simulation begins by assigning an initial odometer value uniformly at random between 0 and 300{,}000~km. 
Each subsequent day, the traveled distance is drawn from a normal distribution with mean \(\mu=200\)~km and standard deviation \(\sigma=20\)~km. 
Maintenance periods influence the signal: no distance is recorded for days entirely within a maintenance interval, while the first and last days of such intervals contribute half of the sampled distance.

At the end of the clean data generation process, we obtain the tuple \(\langle \mathcal{F}, \mathcal{S}, \mathcal{W}, \mathcal{M} \rangle\),  
where \(\mathcal{F}\) denotes the fleet registry, \(\mathcal{S}\) the sensor time-series, \(\mathcal{W}\) the workshop metadata, and \(\mathcal{M}\) the maintenance log.  
The clean maintenance log \(\mathcal{M}\) serves as the ground truth for all downstream evaluation tasks.

\subsubsection*{Noisy Data Sources}

In the noise generation step, the maintenance table \(\mathcal{M}\) is regenerated while preserving the other data sources. 
Each noise generator is invoked in the same sequence as in the clean generation process.  
 
Corruptive noises (M1, M3, M4, M6) operate by copying the clean records $\mathcal{E}$ and modifying selected fields, whereas generative noises (M2, M5) create entirely new records $\mathcal{E}'$ without referencing the clean dataset. 
Specifically, M1 replaces the \texttt{license\_plate} field with another vehicle identifier (e.g., \texttt{device\_table}, \texttt{VIN});  
M3 invalidates service catalog fields (\texttt{system}, \texttt{subsystem}, \texttt{component}, \texttt{activity}) through fixed invalid labels, typos, field swaps, or mismatched hierarchy substitutions;  
M4 clears one or more categorical fields;  
M6 shifts the \texttt{end\_date} field forward, in order to create an inconsistency with the intervention timeline.  
On the generative side, M2 produces valid-looking maintenance entries linked to license plates absent from the monitored fleet, and M5 synthesizes maintenance records documenting testing of the fleet monitoring system.

The noisy-data generation yields the tuple 
\[
\langle \mathcal{F}, \mathcal{S}, \mathcal{W}, \mathcal{M}' \rangle,
\]
where \(\mathcal{M}'\) is the noisy version of the maintenance log, replacing the clean \(\mathcal{M}\) from the noise-free dataset. 
Additionally, it includes, for each record, a label indicating the applied perturbation operator (i.e., the specific noise type) or marking it as \emph{noise-free} when no corruption is present.

\section{Methodology}
\subsection{LLM-empowered log cleaning}

We propose to replace the traditional batch oriented log cleaning process with an LLM empowered stream processing pipeline. 
The key changes lie in two aspects. First, we transition from offline batch processing, where log entries are accumulated and cleaned retrospectively, to real time stream processing, which enables immediate detection and correction of anomalies as records are ingested. 
Second, we augment the stream processing pipeline with a novel LLM-based component. 
This component acts as an intelligent agent that not only detects noisy or incomplete log entries but also performs contextual repairs.

\subsection{Agent Environment}
\label{sec:agent_environment}
\begin{figure}[t]
  \centering
  \includegraphics[width=0.48\textwidth]{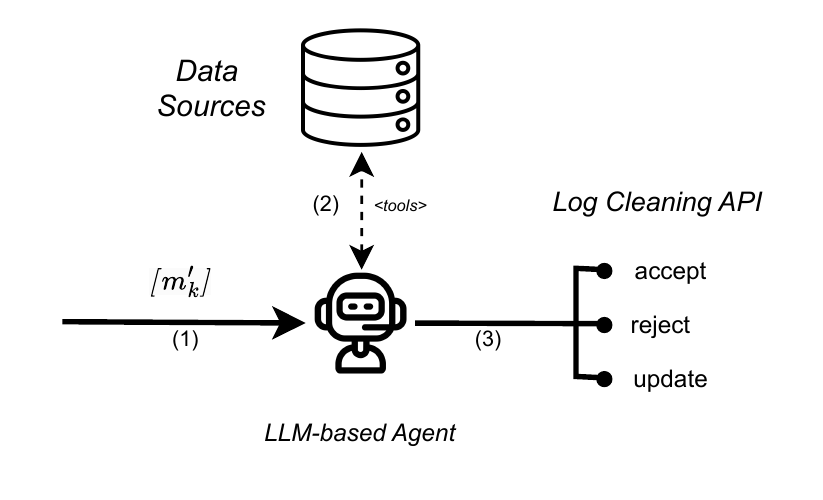}
  \caption{Agent environment with data sources and Log Cleaning API. (1) A noisy record $m'_k$ is provided to the LLM-based agent; (2) the agent optionally queries enterprise data sources through database tools; (3) the agent issues a structured action to the Log Cleaning API: \textit{accept}, \textit{reject}, or \textit{update}.}
  \label{fig:agent_environment}
\end{figure}

To evaluate the ability of LLM-based agents to detect and clean noisy maintenance logs, we design a digital environment that exposes two interfaces to the agent: (i) read-only tools over the enterprise data sources (Section~\ref{sec:data_sources}) and (ii) a dedicated Log Cleaning API. 
The former is instantiated on top of a relational DB while the latter is implemented as a set of agentic output functions.

Table \ref{tab:agent_tools} summarizes the DB tools used in our study.
The agent is equipped with capabilities to list the available database tables, inspect their schema, and query the data they contain. 
In practice, the assumption that all relevant information resides within a single data source does not always hold.
However, in our experimental setup, the focus is not on evaluating the agent’s ability to operate across heterogeneous platforms, but rather on assessing its capacity to successfully complete the curation tasks.

The Log Cleaning API is specified in Table~\ref{tab:log_cleaning_api}.
This API enables the agent to select exactly one of three actions on a record based on its work order number field: \textit{accept}, \textit{reject}, or \textit{update}.

\begin{table}[h]
\centering
\scriptsize
\caption{Database tools available to the agent.}
\label{tab:agent_tools}
\setlength{\tabcolsep}{4pt} 
\begin{tabularx}{\linewidth}{@{}l l Y@{}}
\toprule
\textbf{Tool} & \textbf{Signature} & \textbf{Purpose and Return} \\
\midrule
\texttt{list\_tables} & \texttt{()} &
Enumerates available tables so the agent can discover the database surface before issuing queries. \\
\texttt{describe\_table} & \texttt{(table\_name)} &
Provides schema introspection for a given table (columns and data types). \\
\texttt{run\_sql} & \texttt{(query, limit)} &
Executes a SQL \texttt{SELECT} query with a row cap (\texttt{limit}) to avoid large responses. \\
\bottomrule
\end{tabularx}
\end{table}

\begin{table}[h]
\centering
\scriptsize
\caption{Log Cleaning API methods available to the agent.}
\label{tab:log_cleaning_api}
\setlength{\tabcolsep}{4pt}
\begin{tabularx}{\linewidth}{@{}l l Y@{}}
\toprule
\textbf{Method} & \textbf{Signature} & \textbf{Purpose} \\
\midrule
\texttt{accept} & \texttt{(wo\_num)} &
Confirms the record as clean. \\
\texttt{reject} & \texttt{(wo\_num)} &
Marks the record as out-of-scope or irreparable. \\
\texttt{update} & \texttt{(wo\_num,\ field,\ value)} &
Applies a single field-level correction to the record before accepting it. \\
\bottomrule
\end{tabularx}
\end{table}

\subsection{Task}
The core task is formulated as a three-class classification problem over individual maintenance log entries. 
Given a noisy record m', the LLM-based agent must select one of the following mutually exclusive actions:
\begin{enumerate}
    \item accept - the record is clean and requires no modification.
    \item reject - the record is irreparable or out-of-scope 
    \item update - the record contains correctable noise, and the agent must apply a single-field correction.
\end{enumerate}

We observe that noise-free records must always be accepted, records affected by generative noise  must be rejected, and records affected by corruptive noise must be updated.

Figure~\ref{fig:agent_environment} illustrates the expected end-to-end workflow. 
\textbf{(1)} A raw record $m'_k$ is passed to the LLM-based agent. 
\textbf{(2)} The agent orchestrates a short sequence of tool calls. 
\textbf{(3)} Based on the gathered evidence, the agent must choose exactly one action and submit it to the Log Cleaning API.

The outputs of the experiments fall into four categories. 
In addition to the three valid actions, the task may fail due to the agent's inability to correctly invoke the available tools or output functions. 
These failure cases are explicitly tracked to assess the robustness of each model.

A key constraint in this study is that no examples must be provided in the prompt. 
The agent is guided solely by a system prompt and task instructions, which define its role as a data curator, list the available database tools, and specify the permitted API actions. 
This design choice ensures that the agent must generalize to unseen noise patterns. 
This constraint reflects real-world PdM scenarios, where corrupted records may arrive in isolation, noise patterns may evolve over time, and prior examples may not be available or representative. 
As such, the agent must rely on schema understanding, contextual reasoning, and external data sources to make informed decisions. 

This setup allows us to assess the agent’s robustness and adaptability in realistic, zero-shot conditions.

\begin{table}[t]
\centering
\caption{Selected large language models for benchmarking: context and price (per 1M tokens).}
\label{tab:openrouter_models_requested_column}
\setlength{\tabcolsep}{3pt}
\renewcommand{\arraystretch}{1.08}
\footnotesize
\begin{tabularx}{\columnwidth}{Y l r l}
\toprule
\textbf{Model} & \textbf{Prov.} & \textbf{Ctx (k)} & \textbf{Cost in/out} \\
\midrule
nemotron\mbox{-}nano\mbox{-}9b\mbox{-}v2     & NVIDIA     & 131  & \$0.04 / \$0.16 \\
gpt\mbox{-}oss\mbox{-}20b                     & OpenAI     & 131  & \$0.04 / \$0.15 \\
gpt\mbox{-}oss\mbox{-}120b                    & OpenAI     & 131  & \$0.072 / \$0.28 \\
qwen3\mbox{-}next\mbox{-}80b\mbox{-}a3b\mbox{-}instr. & Qwen       & 262  & \$0.098 / \$0.391 \\
kimi\mbox{-}k2\mbox{-}0905                    & MoonshotAI & 262  & \$0.38 / \$1.522 \\
gpt\mbox{-}5                                  & OpenAI     & 400  & \$1.25 / \$10.00 \\
\bottomrule
\end{tabularx}
\vspace{2pt}
\end{table}

\subsection{Experimental Protocol}

This research aims to evaluate the capability of LLMs to clean noisy maintenance records. 
The evaluation is structured into three main steps: environment generation, prompt engineering, and per-model evaluation.

\paragraph{Environment Generation}
Given a fleet size $N$ and a set of per-noise proportions ${\pi_k}$, we instantiate $R+1$ environments as described in Section~\ref{sec:fleet_data_generation}, using distinct random seeds to ensure statistical independence. 
Each environment is associated with a parameter set $\theta$, sampled randomly from a predefined search space $\Theta$.

\paragraph{Prompt Engineering}
We select an environment and a large language model (outside the evaluation set $\mathcal{LLM}$) and fine-tune a prompt $p$. 
The output of this step is a parametric prompt template $T$, which adapts to each individual record $m$.

\paragraph{Per-Model Evaluation}
For each $llm \in \mathcal{LLM}$ and each environment–parameter pair $(\epsilon, \theta)$, we perform the following steps:

\begin{enumerate}[label=\arabic*)] \item \textbf{Tabular Serialization.}
Flatten the noisy maintenance log of environment $\epsilon$ into an iterable collection
$M_{\text{ser}} = {m^{(i)}}$, where each record is represented as a set of field–value pairs.

\item \textbf{Record Processing.}
For each record $m$ in the serialized maintenance log $M_{\text{ser}}$, we construct a task-specific prompt $p$ tailored to the record's content. 
This prompt is then submitted to the selected language model, along with the environment-specific parameters $\theta$, resulting in an action $a$. 
The action is propagated to the original record to produce a cleaned version $\hat{m}$. 
Each cleaned record is then added to the cumulative set of processed records, denoted by $\hat{M}$.
\end{enumerate}
Finally, we compute evaluation metrics by comparing the cleaned records $\hat{M}$ against the reference targets of environment $env$. 
These metrics are then aggregated across all environments to produce a global performance summary for each model. 
More details about the metrics will be provided in the following sections.

\begin{algorithm}[h!]
\SetAlgoLined
\KwIn{\(LLM, N, \{\pi_k\}, \Theta, R\)}
ENV $\leftarrow$ \{\}\;
\For{$r \leftarrow 1$ \KwTo $R+1$}{
    $env \leftarrow$ GenerateFleetEnv(N, $\{\pi_k\}$)\;
    $ \theta\leftarrow$ SelectParameters($\Theta$)\;
    ENV.append($env$,$\theta$)\;
}
t $\leftarrow$ BuildPromptTemplate(ENV.pop())

\ForEach{$llm \in LLM$}{
    \ForEach{$(env,\theta) \in ENV$}{
        $M'_{ser} \leftarrow $ TabularDataSerialization(env.M')\;
        $\hat{M} \leftarrow$ \{\}\;
        \ForEach{$m' \in M'_{ser}$}{
            $p \leftarrow $ BuildPrompt($t$,$m'$)\;
            $a \leftarrow $ CallLLM($p$, $llm$, $\theta$)\;
            $\hat{m} \leftarrow$ ApplyAction($m'$, $a$)\;
            $\hat{M} \leftarrow \hat{M} \cup {\hat{m}}$.
        }
    }
}
\caption{Experiment Design}
\end{algorithm}

\begin{table*}[t]
\centering
\caption{Error Detection Rate (EDR) and Error Corrected Rate (ECR) by noise type and model}
\label{tab:noise_report_rates}
\renewcommand{\arraystretch}{1.15}
\footnotesize
\begin{tabular}{l cc cc cc cc cc cc}
\toprule
\textbf{Noise} & \multicolumn{2}{c}{\textbf{nemotron}} & \multicolumn{2}{c}{\textbf{gpt-oss-20b}} & \multicolumn{2}{c}{\textbf{kimi-k2}} & \multicolumn{2}{c}{\textbf{qwen3}} & \multicolumn{2}{c}{\textbf{gpt-oss-120b}} & \multicolumn{2}{c}{\textbf{gpt-5}} \\
\cmidrule(lr){2-3} \cmidrule(lr){4-5} \cmidrule(lr){6-7} \cmidrule(lr){8-9} \cmidrule(lr){10-11} \cmidrule(lr){12-13}
 & EDR & ECR & EDR & ECR & EDR & ECR & EDR & ECR & EDR & ECR & EDR & ECR \\
noise free & 96.7\% & -- & 92.7\% & -- & 98.7\% & -- & 92.7\% & -- & 97.3\% & -- & 99.3\% & -- \\
vehicle id mis. & 0.0\% & 0.0\% & 6.7\% & 4.0\% & 6.0\% & 5.3\% & 6.7\% & 2.7\% & 9.3\% & 9.3\% & 27.7\% & 27.7\% \\
out-of-fleet veh.  & 100.0\% & -- & 98.0\% & -- & 96.0\% & -- & 98.0\% & -- & 94.7\% & -- & 98.7\% & -- \\
invalid value & 24.7\% & 21.3\% & 72.7\% & 70.7\% & 82.0\% & 82.0\% & 86.0\% & 86.0\% & 81.3\% & 81.3\% & 83.7\% & 83.7\% \\
missing value & 7.3\% & 0.7\% & 95.3\% & 92.0\% & 81.3\% & 81.3\% & 93.3\% & 87.3\% & 99.3\% & 99.3\% & 100.0\% & 100.0\% \\
digital system test & 98.7\% & -- & 97.3\% & -- & 95.3\% & -- & 56.0\% & -- & 94.7\% & -- & 99.7\% & -- \\
wrong end date & 0.7\% & 0.0\% & 0.7\% & 0.0\% & 0.0\% & 0.0\% & 4.0\% & 0.0\% & 0.0\% & 0.0\% & 0.0\% & 0.0\% \\
\bottomrule
\end{tabular}
\end{table*}

\begin{table*}[t]
\centering
\caption{LLM usage metrics per experiment and model}
\label{tab:llm_usage_metrics}
\footnotesize
\setlength{\tabcolsep}{4.5pt}
\begin{tabular}{l r r r r}
\toprule
\textbf{Model} & \textbf{Request tokens} & \textbf{Response tokens} & \textbf{Time (s)} & \textbf{Cost (USD)} \\
\midrule
\addlinespace[3pt] 
nemotron      & $716250 \pm 14379$   & $393667 \pm 8705$   & $4230.52 \pm 786.67$   & $0.09 \pm 0.00$ \\
gpt-oss-20b   & $1160641 \pm 50397$  & $229529 \pm 18124$  & $3543.66 \pm 423.41$   & $0.08 \pm 0.00$ \\
kimi-k2       & $1921833 \pm 148116$ & $48136 \pm 2736$    & $3008.82 \pm 705.35$   & $0.80 \pm 0.06$ \\
qwen3         & $4944233 \pm 305278$ & $119794 \pm 10829$  & $3213.85 \pm 139.56$   & $0.21 \pm 0.01$ \\
gpt-oss-120b  & $1855565 \pm 48748$  & $169320 \pm 4478$   & $3907.89 \pm 448.30$   & $0.18 \pm 0.00$ \\
gpt-5         & $1043889 \pm 30472$  & $455698 \pm 14705$  & $11051.15 \pm 1718.27$ & $5.86 \pm 0.17$ \\
\bottomrule
\end{tabular}
\end{table*}

\subsection{LLMs}
We evaluate six production LLMs, grouped by capacity into small (Nemotron-Nano-9B-v2), medium (Gpt-Oss-20B), and large (Qwen3-Next-80B-A3B-Instruct, Gpt-Oss-120B, Kimi-K2-0905, and Gpt-5).
These models have been chosen for their agentic capabilities (tool/function calling and schema-constrained outputs), long context windows, and diverse provider ecosystems. 

NVIDIA’s Nemotron is a 9B-parameter hybrid model that combines Mamba-2/MLP layers with a small number of attention blocks, targeting long-context reasoning at modest compute.  
OpenAI’s open-weight Gpt-Oss models use Mixture-of-Experts architecture (MoE). 
Gpt-oss-20b activates ~3.6B parameters per token, while the 120B model activates ~5.1B. 
Qwen3-Next-80B-A3B-Instruct adopts a hybrid MoE layout, with 80B total parameters and ~3B activated per token for efficiency at 256k context. 
Kimi-K2-0905 extends the boundaries of sparse scaling, reporting $\sim$1T total parameters with $\sim$32B active per token and support for 256k-token contexts. 
Finally, Gpt-5 serves as a production baseline at the top end of quality. 
OpenAI does not disclose parameter counts or internal sparsity, so we treat it as a black-box dense system and rely on the public interface for comparability.

Table~\ref{tab:openrouter_models_requested_column} reports the context window sizes and token-level pricing for each large language model selected for benchmarking. These cost estimates are sourced from OpenRouter\footnote{\url{https://openrouter.ai/}}, which provides unified access to a diverse set of commercial and open-weight models through a standardized API. All prices are reported per 1 million tokens, separated into input and output rates, and reflect public pricing tiers at the time of evaluation.

\section{Prompting, Benchmark Configuration and Metrics}

\paragraph{Prompting}

All experiments in this study were performed using zero-shot prompting. The prompt template was manually crafted and evaluated using \emph{Gpt-5 Mini}. The system prompt, user prompt template, and instruction set are provided in the Appendix (see Appendix Figs.~\ref{fig:system_prompt}–\ref{fig:user_prompt}).
\paragraph{Dataset Configuration}

For all experiments, we fix the fleet size to $N = 210$ vehicles and generate a total of $210$ maintenance records per environment. 
The noise distribution is uniform across all categories, with $30$ records for each of the six noise types ($M_1$ to $M_6$) and $30$ noise-free records ($M_0$). 

Despite the fact that real-world maintenance logs exhibit highly skewed noise distributions (e.g., test records occur sporadically, while typos are more frequent), this balanced setup facilitates per-noise-type analysis.

We generate $R = 31$ independent environments using distinct random seeds to ensure statistical robustness. 
Each environment is associated with a pair of LLM-specific decoding parameters, randomly sampled from the space:
\[
\Theta = \left\{ \texttt{temperature} \in (0, 0.2),\ \texttt{top\_p} \in (0.7, 1.0) \right\}
\]
This sampling introduces controlled variability in decoding behavior, allowing us to evaluate model robustness under slight changes in generation dynamics.

\paragraph{Retry Mechanism and Failure Handling}

To account for transient failures and improve robustness, we implement a structured retry mechanism during inference. 
Each record is allowed multiple attempts to be processed successfully before being marked as failed. 
The retry policy is defined as follows:

\begin{itemize}
    \item \textbf{Output generation:} The model is allowed up to 50 retries to produce a valid structured output conforming to the Log Cleaning API schema.
    \item \textbf{Tool invocation:} If the agent fails to execute a tool call (e.g., SQL query or schema inspection), it is allowed up to 3 retries per tool.
    \item \textbf{Record-level recovery:} If a record fails due to persistent output or tool invocation errors, the entire repair process is re-executed up to 3 times before the record is definitively labeled as \texttt{failed}.
\end{itemize}

This mechanism ensures that occasional decoding anomalies or transient tool failures do not disproportionately affect the evaluation metrics. 
All failure cases are explicitly tracked and included in the final analysis to reflect the practical reliability of each model under realistic deployment conditions.

\paragraph{Metrics}

The cleaned dataset $\hat{M}$ is compared against the ground truth to compute:

\begin{itemize}
    \item \textbf{Error Detection Rate (EDR)} — proportion of records for which the correct action was selected.
    \item \textbf{Error Correction Rate (ECR)} — proportion of records for which the correct field-level fix was applied (only for \texttt{update} actions).
\end{itemize}

In addition to task-specific performance, we track the following usage metrics for each experiment execution: total number of request and response tokens, total execution time (in seconds), estimated cost (in USD).

\section{Results}

Table~\ref{tab:noise_report_rates} summarizes performance across six noise categories, as well as the noise-free condition, using the EDR and, where applicable, the ECR. 
Noise-free records were reliably handled by all models: EDRs exceeded 92\% across the board, with \emph{Gpt-5} reaching 99.3\%. 
For generative noise, digital system test entries were almost always flagged correctly. 
Most models achieved near-perfect rejection, and \emph{Gpt-5} led with 99.7\% EDR. 

Corruptive noise showed sharper separation between models. 
Large models handled categorical typos and missing values substantially better than smaller ones: \emph{Gpt-5} attained 83.7\% EDR/ECR on typos and 100\% on missing values, indicating both accurate action selection and successful single-field repair. \textsc{Gpt-Oss-120B} closely tracked this behavior with 81.3\% EDR/ECR on invalid values and 99.3\% on missing values.
In contrast, \emph{Nemotron} struggled on the same categories (24.7\% EDR on typos and 7.3\% on missing values), underscoring sensitivity to fine-grained edits. By comparison, \textsc{Gpt-Oss-20B}, \emph{Kimi-K2}, and \emph{Qwen3} showed better capabilities on these categories.
The two most challenging corruptive cases were wrong end dates and vehicle identifier misalignments, where these models also struggled.
All models failed to correct wrong end dates (0\% ECR), and only \emph{Gpt-5} showed moderate success on identifier misalignment (27.7\% EDR/ECR). 
These results suggest that temporal consistency checks and cross-table identifier resolution remain open problems for current agentic setups.

Table~\ref{tab:llm_usage_metrics} reports resource usage per experiment, revealing a clear cost–quality trade-off. 
\emph{Gpt-5} was the most expensive and slowest configuration (average \$5.86 and 11051\,s per experiment), consistent with its top performance on several categories. 
\emph{Gpt-Oss-120B} offered a favorable price–latency trade-off at roughly 1.86M input and 169k output tokens, 3908\,s runtime, and an estimated \$0.18 per experiment.
\emph{Nemotron}, while the least capable on correction tasks, was the most economical (about \$0.09 per run). 
\emph{Kimi-K2} and \emph{Qwen3} offered a more balanced profile, delivering mid-range EDR/ECR with sub-\$1 costs. 

Overall, higher EDR/ECR on corruptive noise correlates with increased token usage and latency. In contrast, budget-friendly models provide fast, low-cost passes that may suffice for high-confidence acceptance/rejection but lag on precise repairs.

\section{Discussion}

We observed that the agentic approach enabled a shift from batch processing to stream processing in maintenance log cleaning. The methodology simulated stream processing by handling one record at a time, which—through integration with external tools—enabled intelligent and context-aware data curation. A key advantage was the autonomy of the agents: they operated without explicit cleaning task specification. The agents also demonstrated contextual reasoning by leveraging multiple tables in real time when inferring corrections. Moreover, this approach enabled the extension of data curation tasks to include time-series information, whereas traditional solutions were typically limited to tabular formats.

From a cost–quality perspective, among all models \textsc{Gpt-Oss-120B} stands out as a strong value option. It delivered a favorable price–latency profile while remaining competitive on generic repairs. These characteristics make it attractive when budgets or latency constraints are tight, especially relative to the premium \textsc{Gpt-5} configuration. Moreover, because the \textsc{Gpt-Oss} line is open-weight, \textsc{Gpt-Oss-120B} can be executed and fine-tuned on private infrastructure to target domain-specific noise patterns.

Notably, even small and medium-sized models demonstrated the ability to perform cleaning tasks. For instance, \emph{Nemotron}, with only 9B parameters, successfully detected generative noise, while \emph{Gpt-Oss-20B} handled invalid and missing values with a good degree of accuracy. 

The response time for processing individual records ranged from a few seconds to several minutes, which we consider acceptable for small and medium fleet contexts, given the rarity of failures.

Despite these promising results in performing sector-agnostic data cleaning, the inability to handle domain-specific noise remained a significant limitation. Errors detectable through temporal misalignments and entity association are precisely where LLM agents could provide the most value. We believe that current LLMs have not been trained on structured, domain-specific cases and therefore lack the inductive bias required to generalize effectively in such contexts. Consequently, more advanced prompting strategies and fine-tuning, though at the expense of autonomy, could be employed to improve performance in these areas.

Finally, while the synthetic-data generator enabled controlled benchmarking and reproducibility, reliance on synthetic logs limits the external validity of our findings in operational settings. This choice reflects privacy and confidentiality constraints around proprietary fleet data. To close this gap, it is necessary to deploy and evaluate the pipeline on de-identified maintenance logs from industry.

\section{Conclusions and Future Work}

In this study, we investigated the potential of LLM agents to clean noisy maintenance logs in PdM applications. 
We introduced a synthetic data generation framework that simulates realistic noise patterns across six categories, including both generic and domain-specific anomalies. 
We benchmarked six production-grade LLMs using a stream-based agentic setup, where each model was tasked with classifying and repairing individual log entries via structured API calls. 
Performance was evaluated using error detection rate and error correction rate, alongside usage metrics such as runtime, token consumption, and cost. LLM agents performed well at identifying corruptive noise, recognizing noise-free records, and carrying out domain-agnostic repairs. 
By contrast, they underperformed on domain-specific noise patterns, where schema and process knowledge are required. 
These findings should be interpreted in light of our use of synthetic logs, which may limit external validity.

Building on this initial investigation, several extensions are envisioned to further advance the evaluation and the applicability of LLM agents in automotive industrial settings. 
First, the synthetic fleet data generator should incorporate an expanded noise taxonomy that captures more complex errors, such as multi-field corruptions, time-series inconsistencies, inter-record contradictions, and semantic mismatches. 
Additionally, the data schema could be enriched with nested structures and optional fields to better capture the complexity and variability of real-world maintenance logs. To address the current limitations in handling temporal inconsistencies and cross-table reasoning, we propose three key enhancements: (i) integrating temporal-logic validators into the agent’s toolset to enable consistency checks across time-stamped fields; (ii) adopting a hybrid rule–LLM architecture, where deterministic rules are used to pre-validate and post-validate temporal fields and entity relationships, while the LLM focuses on proposing semantic repairs; and (iii) fine-tuning the models on domain-specific corpora to improve their inductive bias and contextual understanding. Furthermore, we plan to incorporate persistent memory, allowing agents to maintain context across multiple records and leverage historical decisions for more coherent and informed reasoning.

Ultimately, to strengthen external validity and enhance the practical significance of our findings, we plan to evaluate the agents using authentic maintenance logs obtained from industrial partners, with appropriate anonymization protocols in place. This will allow us to rigorously assess the generalizability, robustness, and domain-specific adaptability of the proposed LLM-based cleaning approach under real-world conditions.

Overall, LLM-based maintenance log cleaning shows strong potential to outperform traditional approaches in terms of autonomy, flexibility, and real-time responsiveness. While current limitations exist, we believe these can be mitigated. As such, LLMs represent a promising direction for future PdM data cleaning pipelines, especially as the technology continues to mature.

\section*{Acknowledgement}

\label{sec:acknowledgment}
This research was funded in whole, or in part, by the Luxembourg National Research Fund (FNR), grant reference BRIDGES/2022/IS/17270233. For the purpose of open access, and in fulfillment of the obligations arising from the grant agreement, the authors have applied a Creative Commons Attribution 4.0 International (CC BY 4.0) license to any Author Accepted Manuscript version arising from this submission.

The authors gratefully acknowledge Grupo\textless A\textgreater ~for providing access to the dataset used in this study, which was essential for the development and validation of the research.


\begin{thebibliography}{}

\bibitem [\protect \citeauthoryear {%
Bendinelli%
, Dox%
\BCBL {}\ \BBA {} Holz%
}{%
Bendinelli%
\ \protect \BOthers {.}}{%
{\protect \APACyear {2025}}%
}]{%
bendinelli_exploring_2025}
\APACinsertmetastar {%
bendinelli_exploring_2025}%
\begin{APACrefauthors}%
Bendinelli, T.%
, Dox, A.%
\BCBL {}\ \BBA {} Holz, C.%
\end{APACrefauthors}%
\unskip\
\newblock
\APACrefYearMonthDay{2025}{}{}.
\newblock
\APACrefbtitle {Exploring {LLM} Agents for Cleaning Tabular Machine Learning Datasets} {Exploring {LLM} agents for cleaning tabular machine learning datasets}\ (\BNUM\ {arXiv}:2503.06664).
\newblock
\APACaddressPublisher{}{{arXiv}}.
\newblock
\begin{APACrefDOI} \doi{10.48550/arXiv.2503.06664} \end{APACrefDOI}
\PrintBackRefs{\CurrentBib}

\bibitem [\protect \citeauthoryear {%
Chu%
\ \protect \BOthers {.}}{%
Chu%
\ \protect \BOthers {.}}{%
{\protect \APACyear {2015}}%
}]{%
chu_katara_2015}
\APACinsertmetastar {%
chu_katara_2015}%
\begin{APACrefauthors}%
Chu, X.%
, Morcos, J.%
, Ilyas, I\BPBI F.%
, Ouzzani, M.%
, Papotti, P.%
, Tang, N.%
\BCBL {}\ \BBA {} Ye, Y.%
\end{APACrefauthors}%
\unskip\
\newblock
\APACrefYearMonthDay{2015}{}{}.
\newblock
{\BBOQ}\APACrefatitle {{KATARA}: A Data Cleaning System Powered by Knowledge Bases and Crowdsourcing} {{KATARA}: A data cleaning system powered by knowledge bases and crowdsourcing}.{\BBCQ}
\newblock
\BIn{} \APACrefbtitle {Proceedings of the 2015 {ACM} {SIGMOD} International Conference on Management of Data} {Proceedings of the 2015 {ACM} {SIGMOD} international conference on management of data}\ (\BPGS\ 1247--1261).
\newblock
\APACaddressPublisher{}{{ACM}}.
\newblock
\begin{APACrefDOI} \doi{10.1145/2723372.2749431} \end{APACrefDOI}
\PrintBackRefs{\CurrentBib}

\bibitem [\protect \citeauthoryear {%
Del~Moral%
, Nowaczyk%
\BCBL {}\ \BBA {} Pashami%
}{%
Del~Moral%
\ \protect \BOthers {.}}{%
{\protect \APACyear {2022}}%
}]{%
delmoral2022}
\APACinsertmetastar {%
delmoral2022}%
\begin{APACrefauthors}%
Del~Moral, P.%
, Nowaczyk, S.%
\BCBL {}\ \BBA {} Pashami, S.%
\end{APACrefauthors}%
\unskip\
\newblock
\APACrefYearMonthDay{2022}{}{}.
\newblock
{\BBOQ}\APACrefatitle {Filtering Misleading Repair Log Labels to Improve Predictive Maintenance Models} {Filtering misleading repair log labels to improve predictive maintenance models}.{\BBCQ}
\newblock
\BIn{} \APACrefbtitle {Proceedings of the European Conference of the PHM Society 2022} {Proceedings of the european conference of the phm society 2022}\ (\BVOL~7, \BPGS\ 110--117).
\newblock
\begin{APACrefDOI} \doi{10.36001/phme.2022.v7i1.3360} \end{APACrefDOI}
\PrintBackRefs{\CurrentBib}

\bibitem [\protect \citeauthoryear {%
Fan%
\ \BBA {} Geerts%
}{%
Fan%
\ \BBA {} Geerts%
}{%
{\protect \APACyear {2012}}%
}]{%
fan_foundations_2012}
\APACinsertmetastar {%
fan_foundations_2012}%
\begin{APACrefauthors}%
Fan, W.%
\BCBT {}\ \BBA {} Geerts, F.%
\end{APACrefauthors}%
\unskip\
\newblock
\APACrefYear{2012}.
\newblock
\APACrefbtitle {Foundations of Data Quality Management} {Foundations of data quality management}\ (\BVOL~4).
\newblock
\APACaddressPublisher{}{Morgan and Claypool}.
\newblock
\begin{APACrefDOI} \doi{10.2200/S00439ED1V01Y201207DTM030} \end{APACrefDOI}
\PrintBackRefs{\CurrentBib}

\bibitem [\protect \citeauthoryear {%
Heidari%
, McGrath%
, Ilyas%
\BCBL {}\ \BBA {} Rekatsinas%
}{%
Heidari%
\ \protect \BOthers {.}}{%
{\protect \APACyear {2019}}%
}]{%
heidari_holodetect_2019}
\APACinsertmetastar {%
heidari_holodetect_2019}%
\begin{APACrefauthors}%
Heidari, A.%
, McGrath, J.%
, Ilyas, I\BPBI F.%
\BCBL {}\ \BBA {} Rekatsinas, T.%
\end{APACrefauthors}%
\unskip\
\newblock
\APACrefYearMonthDay{2019}{}{}.
\newblock
{\BBOQ}\APACrefatitle {{HoloDetect}: Few-Shot Learning for Error Detection} {{HoloDetect}: Few-shot learning for error detection}.{\BBCQ}.
\newblock
\begin{APACrefURL} [{2025-08-13}]\url{http://arxiv.org/abs/1904.02285} \end{APACrefURL}
\newblock
\begin{APACrefDOI} \doi{10.1145/3299869.3319888} \end{APACrefDOI}
\PrintBackRefs{\CurrentBib}

\bibitem [\protect \citeauthoryear {%
Ilyas%
\ \BBA {} Chu%
}{%
Ilyas%
\ \BBA {} Chu%
}{%
{\protect \APACyear {2015}}%
}]{%
ilyas_trends_2015}
\APACinsertmetastar {%
ilyas_trends_2015}%
\begin{APACrefauthors}%
Ilyas, I\BPBI F.%
\BCBT {}\ \BBA {} Chu, X.%
\end{APACrefauthors}%
\unskip\
\newblock
\APACrefYearMonthDay{2015}{}{}.
\newblock
{\BBOQ}\APACrefatitle {Trends in Cleaning Relational Data: Consistency and Deduplication} {Trends in cleaning relational data: Consistency and deduplication}.{\BBCQ}
\newblock
\APACjournalVolNumPages{}{5}{4}{281--393}.
\newblock
\begin{APACrefDOI} \doi{10.1561/1900000045} \end{APACrefDOI}
\PrintBackRefs{\CurrentBib}

\bibitem [\protect \citeauthoryear {%
Madhikermi%
, Buda%
, Dave%
\BCBL {}\ \BBA {} Framling%
}{%
Madhikermi%
\ \protect \BOthers {.}}{%
{\protect \APACyear {2017}}%
}]{%
madhikermi_key_2017}
\APACinsertmetastar {%
madhikermi_key_2017}%
\begin{APACrefauthors}%
Madhikermi, M.%
, Buda, A.%
, Dave, B.%
\BCBL {}\ \BBA {} Framling, K.%
\end{APACrefauthors}%
\unskip\
\newblock
\APACrefYearMonthDay{2017}{}{}.
\newblock
{\BBOQ}\APACrefatitle {Key data quality pitfalls for condition based maintenance} {Key data quality pitfalls for condition based maintenance}.{\BBCQ}
\newblock
\BIn{} \APACrefbtitle {2017 2nd International Conference on System Reliability and Safety ({ICSRS})} {2017 2nd international conference on system reliability and safety ({ICSRS})}\ (\BPGS\ 474--480).
\newblock
\APACaddressPublisher{}{{IEEE}}.
\newblock
\begin{APACrefDOI} \doi{10.1109/ICSRS.2017.8272868} \end{APACrefDOI}
\PrintBackRefs{\CurrentBib}

\bibitem [\protect \citeauthoryear {%
Mahdavi%
\ \BBA {} Abedjan%
}{%
Mahdavi%
\ \BBA {} Abedjan%
}{%
{\protect \APACyear {2020}}%
}]{%
mahdavi_baran_2020}
\APACinsertmetastar {%
mahdavi_baran_2020}%
\begin{APACrefauthors}%
Mahdavi, M.%
\BCBT {}\ \BBA {} Abedjan, Z.%
\end{APACrefauthors}%
\unskip\
\newblock
\APACrefYearMonthDay{2020}{}{}.
\newblock
{\BBOQ}\APACrefatitle {Baran: effective error correction via a unified context representation and transfer learning} {Baran: effective error correction via a unified context representation and transfer learning}.{\BBCQ}
\newblock
\APACjournalVolNumPages{}{13}{12}{1948--1961}.
\newblock
\begin{APACrefDOI} \doi{10.14778/3407790.3407801} \end{APACrefDOI}
\PrintBackRefs{\CurrentBib}

\bibitem [\protect \citeauthoryear {%
Mahdavi%
\ \protect \BOthers {.}}{%
Mahdavi%
\ \protect \BOthers {.}}{%
{\protect \APACyear {2019}}%
}]{%
mahdavi_raha_2019}
\APACinsertmetastar {%
mahdavi_raha_2019}%
\begin{APACrefauthors}%
Mahdavi, M.%
, Abedjan, Z.%
, Castro~Fernandez, R.%
, Madden, S.%
, Ouzzani, M.%
, Stonebraker, M.%
\BCBL {}\ \BBA {} Tang, N.%
\end{APACrefauthors}%
\unskip\
\newblock
\APACrefYearMonthDay{2019}{}{}.
\newblock
{\BBOQ}\APACrefatitle {Raha: A Configuration-Free Error Detection System} {Raha: A configuration-free error detection system}.{\BBCQ}
\newblock
\BIn{} \APACrefbtitle {Proceedings of the 2019 International Conference on Management of Data} {Proceedings of the 2019 international conference on management of data}\ (\BPGS\ 865--882).
\newblock
\APACaddressPublisher{}{{ACM}}.
\newblock
\begin{APACrefDOI} \doi{10.1145/3299869.3324956} \end{APACrefDOI}
\PrintBackRefs{\CurrentBib}

\bibitem [\protect \citeauthoryear {%
Narayan%
, Chami%
, Orr%
\BCBL {}\ \BBA {} Ré%
}{%
Narayan%
\ \protect \BOthers {.}}{%
{\protect \APACyear {2022}}%
}]{%
narayan_can_2022}
\APACinsertmetastar {%
narayan_can_2022}%
\begin{APACrefauthors}%
Narayan, A.%
, Chami, I.%
, Orr, L.%
\BCBL {}\ \BBA {} Ré, C.%
\end{APACrefauthors}%
\unskip\
\newblock
\APACrefYearMonthDay{2022}{}{}.
\newblock
{\BBOQ}\APACrefatitle {Can Foundation Models Wrangle Your Data?} {Can foundation models wrangle your data?}{\BBCQ}
\newblock
\APACjournalVolNumPages{}{16}{4}{738--746}.
\newblock
\begin{APACrefDOI} \doi{10.14778/3574245.3574258} \end{APACrefDOI}
\PrintBackRefs{\CurrentBib}

\bibitem [\protect \citeauthoryear {%
Prytz%
, Nowaczyk%
, Rognvaldsson%
\BCBL {}\ \BBA {} Byttner%
}{%
Prytz%
\ \protect \BOthers {.}}{%
{\protect \APACyear {2015}}%
}]{%
prytz2015}
\APACinsertmetastar {%
prytz2015}%
\begin{APACrefauthors}%
Prytz, R.%
, Nowaczyk, S.%
, Rognvaldsson, T.%
\BCBL {}\ \BBA {} Byttner, S.%
\end{APACrefauthors}%
\unskip\
\newblock
\APACrefYearMonthDay{2015}{}{}.
\newblock
{\BBOQ}\APACrefatitle {Predicting the need for vehicle compressor repairs using maintenance records and logged vehicle data} {Predicting the need for vehicle compressor repairs using maintenance records and logged vehicle data}.{\BBCQ}
\newblock
\APACjournalVolNumPages{Engineering Applications of Artificial Intelligence}{41}{}{139--150}.
\newblock
\begin{APACrefDOI} \doi{10.1016/j.engappai.2015.02.009} \end{APACrefDOI}
\PrintBackRefs{\CurrentBib}

\bibitem [\protect \citeauthoryear {%
Qi%
, Miao%
\BCBL {}\ \BBA {} Wang%
}{%
Qi%
\ \protect \BOthers {.}}{%
{\protect \APACyear {2025}}%
}]{%
qi_cleanagent_2025}
\APACinsertmetastar {%
qi_cleanagent_2025}%
\begin{APACrefauthors}%
Qi, D.%
, Miao, Z.%
\BCBL {}\ \BBA {} Wang, J.%
\end{APACrefauthors}%
\unskip\
\newblock
\APACrefYearMonthDay{2025}{}{}.
\newblock
\APACrefbtitle {{CleanAgent}: Automating Data Standardization with {LLM}-based Agents} {{CleanAgent}: Automating data standardization with {LLM}-based agents}\ (\BNUM\ {arXiv}:2403.08291).
\newblock
\APACaddressPublisher{}{{arXiv}}.
\newblock
\begin{APACrefDOI} \doi{10.48550/arXiv.2403.08291} \end{APACrefDOI}
\PrintBackRefs{\CurrentBib}

\bibitem [\protect \citeauthoryear {%
Rekatsinas%
, Chu%
, Ilyas%
\BCBL {}\ \BBA {} Ré%
}{%
Rekatsinas%
\ \protect \BOthers {.}}{%
{\protect \APACyear {2017}}%
}]{%
rekatsinas_holoclean_2017}
\APACinsertmetastar {%
rekatsinas_holoclean_2017}%
\begin{APACrefauthors}%
Rekatsinas, T.%
, Chu, X.%
, Ilyas, I\BPBI F.%
\BCBL {}\ \BBA {} Ré, C.%
\end{APACrefauthors}%
\unskip\
\newblock
\APACrefYearMonthDay{2017}{}{}.
\newblock
{\BBOQ}\APACrefatitle {{HoloClean}: holistic data repairs with probabilistic inference} {{HoloClean}: holistic data repairs with probabilistic inference}.{\BBCQ}
\newblock
\APACjournalVolNumPages{}{10}{11}{1190--1201}.
\newblock
\begin{APACrefDOI} \doi{10.14778/3137628.3137631} \end{APACrefDOI}
\PrintBackRefs{\CurrentBib}

\bibitem [\protect \citeauthoryear {%
Woods%
, Selway%
, Bikaun%
, Stumptner%
\BCBL {}\ \BBA {} Hodkiewicz%
}{%
Woods%
\ \protect \BOthers {.}}{%
{\protect \APACyear {2024}}%
}]{%
woods_ontology_2024}
\APACinsertmetastar {%
woods_ontology_2024}%
\begin{APACrefauthors}%
Woods, C.%
, Selway, M.%
, Bikaun, T.%
, Stumptner, M.%
\BCBL {}\ \BBA {} Hodkiewicz, M.%
\end{APACrefauthors}%
\unskip\
\newblock
\APACrefYearMonthDay{2024}{}{}.
\newblock
{\BBOQ}\APACrefatitle {An ontology for maintenance activities and its application to data quality} {An ontology for maintenance activities and its application to data quality}.{\BBCQ}
\newblock
\APACjournalVolNumPages{}{15}{2}{319--352}.
\newblock
\begin{APACrefDOI} \doi{10.3233/SW-233299} \end{APACrefDOI}
\PrintBackRefs{\CurrentBib}

\bibitem [\protect \citeauthoryear {%
Zhang%
, Dong%
, Xiao%
\BCBL {}\ \BBA {} Oyamada%
}{%
Zhang%
\ \protect \BOthers {.}}{%
{\protect \APACyear {2024}}%
}]{%
zhang_large_2024}
\APACinsertmetastar {%
zhang_large_2024}%
\begin{APACrefauthors}%
Zhang, H.%
, Dong, Y.%
, Xiao, C.%
\BCBL {}\ \BBA {} Oyamada, M.%
\end{APACrefauthors}%
\unskip\
\newblock
\APACrefYearMonthDay{2024}{}{}.
\newblock
\APACrefbtitle {Large Language Models as Data Preprocessors} {Large language models as data preprocessors}\ (\BNUM\ {arXiv}:2308.16361).
\newblock
\APACaddressPublisher{}{{arXiv}}.
\newblock
\begin{APACrefDOI} \doi{10.48550/arXiv.2308.16361} \end{APACrefDOI}
\PrintBackRefs{\CurrentBib}

\end{thebibliography}

\section*{Appendix}
\begin{figure}[h]
\centering
\begin{minipage}{0.48\textwidth}
\begin{promptbox}{System Prompt}

\small
You are a meticulous data curator focused on workshop maintenance logs.\\

When correcting records, you access the following resources in the database:\\
\begin{itemize}[leftmargin=*, itemsep=2pt]
  \item \texttt{fleet\_registry} — contains all vehicles belonging to the target fleet.
  \item \texttt{service\_catalog} — defines valid categories and their hierarchical structure for maintenance records.
  \item \texttt{signal\_odometer} — tracks odometer readings for all vehicles in the fleet.\\
\end{itemize}
Use these resources to validate and correct incoming maintenance records.
\end{promptbox}
\end{minipage}\hfill
\caption{System Prompt.}
\label{fig:system_prompt}
\end{figure}

\begin{figure}[t]
\centering
\begin{minipage}{0.48\textwidth}
\begin{promptbox}{Instructions}
\small
Spot inconsistencies, check DB tables and propose corrections when needed.
\end{promptbox}
\end{minipage}
\caption{Agent instructions.}
\label{fig:instructions}
\end{figure}

\begin{figure}[h]
\centering
\begin{minipage}{0.48\textwidth}
\begin{promptbox}{User Prompt Template}
\small
You are given the next maintenance record:

\begin{center}
\(\langle \text{\textit{record}} \rangle\)
\end{center}

Select exactly one action and invoke the corresponding output function:\\
\begin{itemize}[leftmargin=*, itemsep=2pt]
  \item \texttt{accept(work\_order\_number)} — the record is valid and requires no changes.
  \item \texttt{reject(work\_order\_number)} — the record is invalid or out of scope for this fleet.
  \item \texttt{update(work\_order\_number, field, value)} — the record contains a correctable error; apply a single-field fix.\\
\end{itemize}
Use the appropriate function to classify the record.
\end{promptbox}
\end{minipage}\hfill
\caption{User prompt template used to guide agentic decision-making.}
\label{fig:user_prompt}
\end{figure}

\end{document}